%% file: main.tex
\documentclass{article}
\usepackage{spconf,amsmath,graphicx}
\usepackage{multirow, array, float, dblfloatfix}

\usepackage{enumitem}
\setlist{nosep, leftmargin=14pt}

\title{Chronic Kidney Disease Prognosis Prediction Using Transformer}

\usepackage[hang, flushmargin]{footmisc}
\name{Yohan Lee$^{1,*}$, Dong Gyun Kang$^{1,4,*}$, Sehoon Park$^{3}$, Kwangsoo Kim$^{1,\dagger}$, Sa-Yoon Park$^{2,\dagger}$
\thanks{$*$These authors contributed eequally to this work. \\$\dagger$Corresponding authors.}}
\address{
$^{1}$Department of Transdisciplinary Medicine, Seoul National University Hospital, Seoul, South Korea \\ 
$^{2}$Department of Physiology, College of Korean Medicine, Wonkwang University, Iksan, South Korea \\
$^{3}$Department of Nephrology, Seoul National University Hospital, Seoul, South Korea\\
$^{4}$Department of Computer Science, Georgia Institute of Technology, Atlanta, GA, USA\\
}

\begin{document}
\maketitle

\begin{abstract}
Chronic Kidney Disease (CKD) affects nearly 10\% of the global population and often progresses to end-stage renal failure. Accurate prognosis prediction is vital for timely interventions and resource optimization. We present a transformer-based framework for predicting CKD progression using multi-modal electronic health records (EHR) from the Seoul National University Hospital OMOP Common Data Model. Our approach (\textbf{ProQ-BERT}) integrates demographic, clinical, and laboratory data, employing quantization-based tokenization for continuous lab values and attention mechanisms for interpretability. The model was pretrained with masked language modeling and fine-tuned for binary classification tasks predicting progression from stage 3a to stage 5 across varying follow-up and assessment periods. Evaluated on a cohort of 91,816 patients, our model consistently outperformed CEHR-BERT, achieving ROC-AUC up to 0.995 and PR-AUC up to 0.989 for short-term prediction. These results highlight the effectiveness of transformer architectures and temporal design choices in clinical prognosis modeling, offering a promising direction for personalized CKD care.
\end{abstract}


\input{1_introduction}
\input{2_methods}
\input{3_experiments}
\input{4_results}
\input{5_conclusion}

\clearpage

\section{Compliance with ethical standards}
\label{sec:ethics}
This retrospective study using was approved by the Institutional Review Board of Seoul National University Hospital (Seoul, South Korea)

\section{Acknowledgments}
\label{sec:acknowledgments}
This research was supported by Basic Science Research Program through the National Research Foundation of Korea (NRF) funded by the Ministry of Education (RS-2023-00248152)

\bibliographystyle{IEEEbib}
\bibliography{refs}

\end{document}

%% file: 1_introduction.tex
\section{Introduction}
\label{sec:intro}

\textbf{Chronic kidney disease (CKD)} is a progressive condition affecting millions worldwide, up to 10\% of the global population, often leading to end-stage renal failure and significant morbidity \cite{Jadoul2024}. Accurate prognosis prediction is critical for guiding treatment strategies, optimizing resource allocation, and improving patient outcomes. Traditional prognostic models rely on statistical approaches or shallow machine learning techniques, which often struggle to capture the complex temporal and contextual relationships inherent in longitudinal clinical data.

Recent advances in deep learning, particularly transformer architectures\cite{AttAllNeed}, have revolutionized sequence modeling by leveraging self-attention mechanisms to capture long-range dependencies and contextual patterns. Unlike recurrent models like Recurrent Neural Network (RNN), transformers enable parallel processing and dynamic weighting of features, making them well-suited for heterogeneous healthcare data such as electronic health records, laboratory results, and demographic information. 

Previous studies on prognosis prediction of chronic kidney disease using deep learning have primarily employed recurrent neural networks (RNNs) \cite{Zhu2023} and transformer-based architectures \cite{Li2020BEHRT}\cite{awan-etal-2025-predicting}\cite{pang_cehr-bert_nodate}. However, these approaches have not fully explored the integration of transformer models with laboratory values. Recently, foundational models have attempted to incorporate laboratory data, as demonstrated by Ethos \cite{renc_zero_2024}, which discretized continuous lab values into 10 quantiles for model input. Building on this idea, we adapted the decoder-based architecture of Ethos into an encoder-based model, achieving significant performance improvements in prognosis prediction.

In this study, we propose \textbf{Pro}gnosis Prediction with \textbf{Q}uantization using \textbf{BERT} (\textbf{ProQ-BERT}), a transformer-based framework for predicting CKD prognosis, aiming to enhance predictive accuracy and interpretability compared to conventional methods. Our approach integrates quantification of laboratory data to identify key clinical factors influencing chronic kidney disease progression.

%% file: 2_methods.tex
\section{Methods}
\label{sec:methods}

\subsection{Data}
We use the Seoul National University Hospital Common Data Model (SNUH CDM), which is based on the OMOP CDM (Observational Medical Outcomes Partnership Common Data Model), as our primary data source. It contains de-identified records of more than 3.7 million patients spanning 20 years from 2004 to 2024.
We use six domain tables: \textit{person}, \textit{visit\_occurrence}, \textit{condition\_occurrence}, \textit{drug\_exposure}, \textit{procedure\_occurrence}, and \textit{measurement}. From \textit{person}, we extract person ID, gender, and birth year. From \textit{visit\_occurrence}, we use person ID, visit ID, visit type (outpatient, inpatient, or emergency), and start/end datetime. From \textit{condition\_occurrence}, \textit{drug\_exposure}, and \textit{procedure\_occurrence}, we use their respective event IDs, concept IDs, and datetime, along with person and visit IDs for linkage. From \textit{measurement}, which contains lab test results, we use measurement ID, concept ID, datetime, and numerical value columns. We explain how we utilize these numerical values in the following subsection.

\subsection{Study Design}
\subsubsection{Cohort Definition}
    \begin{enumerate}
    \item \textbf{Inclusion Criteria}
    \begin{enumerate}
        \item Diagnosed with chronic kidney disease (CKD) or end-stage renal disease (ESRD) based on OMOP CDM concept\_id.
        \item Estimated glomerular filtration rate (eGFR) below $60\ \text{mL/min/1.73 m}^2$ continuously for at least 90 days
        \item Urine albumin-to-creatinine ratio (uACR) above $30\ \text{mg/g}$ continuously for at least 90 days
    \end{enumerate}
    
    \vspace{0.25cm}
    
        \item \textbf{Exclusion Criteria}
        \begin{enumerate}
            \item Patients who never measured estimated glomerular filtration rate (eGFR) in Seoul National University Hospital (SNUH)
        \end{enumerate}
    \end{enumerate}
    
    \vspace{0.25cm}

\begin{figure}[t]
\centering
\includegraphics[width=\columnwidth]{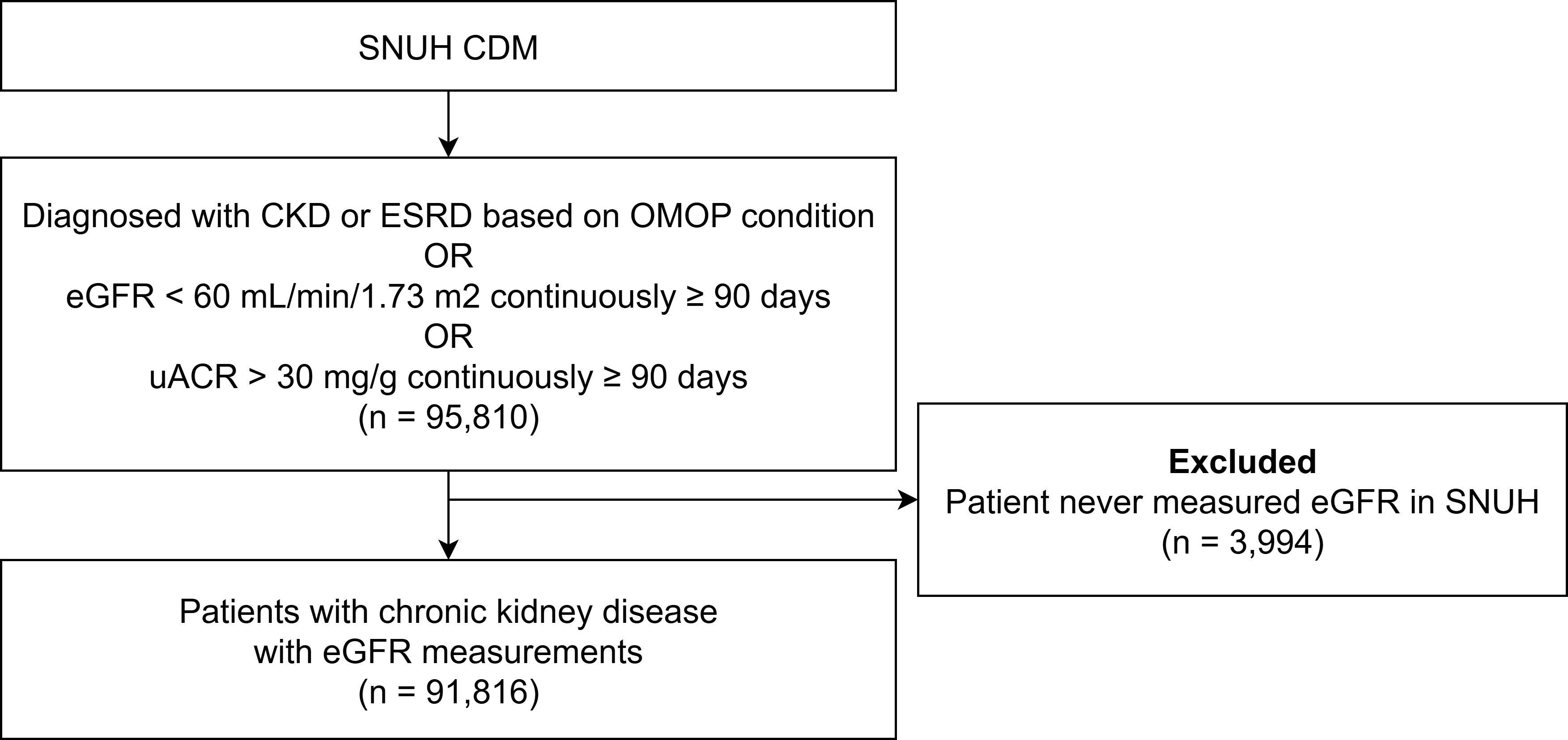}
\caption{Cohort Definition Flowchart}
\label{fig:cohort_def}
\end{figure}

The cohort used in this study was defined by the inclusion and exclusion criteria above. A patient was included if they met any of the inclusion criteria. This resulted in a total of 91,816 patients.

\subsubsection{Definition of Outcome}
    We defined the outcome based on the progression of chronic kidney disease (CKD) stages. For each patient, an index date was assigned for CKD stage 3a and stage 5. The threshold for stage 3a was an estimated glomerular filtration rate (eGFR) of $< 60 mL/min/1.73 m^2$, and for stage 5, $< 15 mL/min/1.73 m^2$. The index date for each stage was determined as the first date on which the eGFR fell below the respective threshold and persisted for at least 90 days. 
    
    The primary outcome, worsening prognosis, was defined as the occurrence of stage 5 within a specified follow-up period after the stage 3a index date. Patients were classified as:
    
    Cases: Those who reached stage 5 within the follow-up period after the stage 3a index date.
    
    Controls: Those with follow-up data exceeding the defined follow-up period and without progression to stage 5 during that time.
    
    The predictive model was trained to classify patients as cases or controls using only data available within the assessment period preceding the stage 3a index date, ensuring no information leakage from the follow-up period. Multiple follow-up durations and assessment periods were evaluated to examine the robustness of the findings.

\begin{figure*}[t]
\centering
\includegraphics[width=\textwidth]{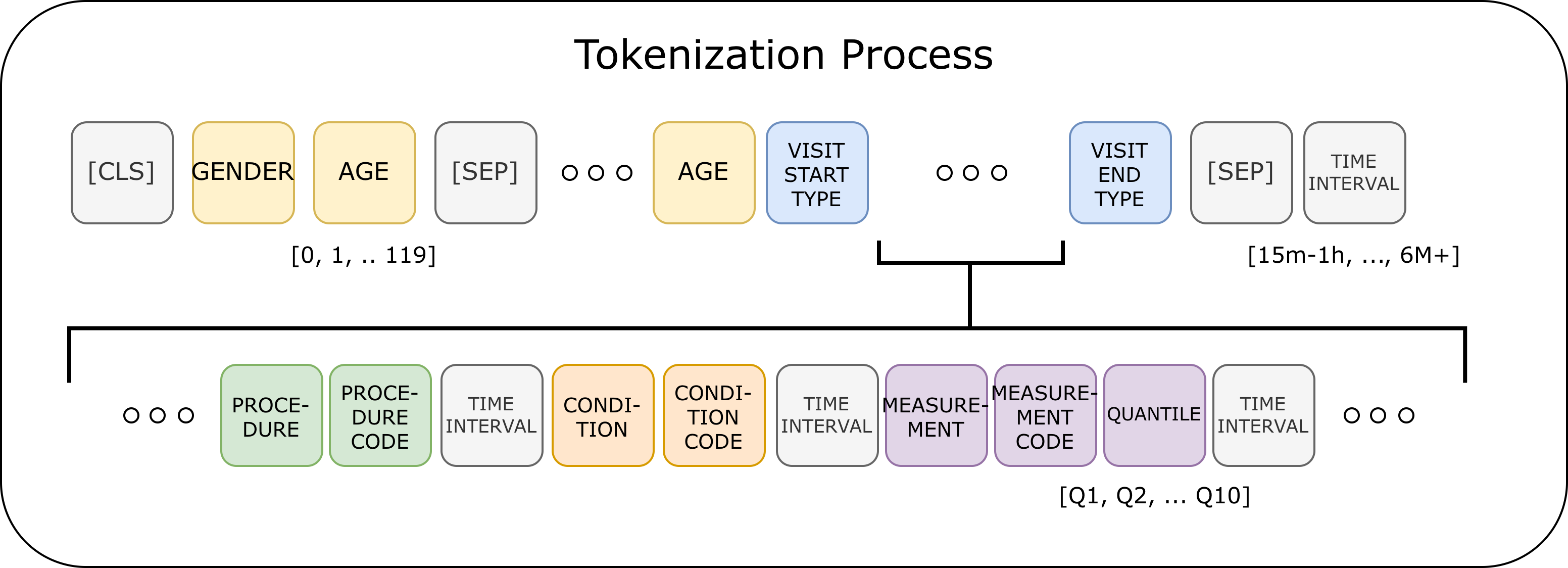}
\caption{Patient sequence tokenization process showing demographic tokens, visit segments, medical events with temporal relationships, and quantized laboratory measurements.}
\label{fig:tokenization}
\end{figure*}

\subsection{Patient Sequences and Tokenization}
As depicted in Figure~\ref{fig:tokenization}, the core of the method lies in aggregating patient data from various CDM domains and converting them into sequences representing the patients' medical events. To achieve this, we first gather each patient's medical events in chronological order and build sequences from the gathered events. A sequence contains all medical events of a patient for a designated period, where each medical event is converted to a token to compose the patient's sequence, along with other tokens that function as special tokens or represent patient static information.

Each sequence starts with a [CLS] token, followed by the patient's gender token. [SEP] tokens are used as delimiters between temporal segments. Medical events are converted to tokens according to their unique concept IDs, placed after their corresponding type tokens ([CONDITION], [DRUG], [PROCEDURE], [LAB]). Time interval tokens (e.g., [TIME\_5m-15m], [TIME\_1h-2h]) are inserted between consecutive events to capture temporal relationships.

Medical events such as procedures, drugs, and conditions occur within visits, so they are positioned between visit start and end event tokens. We assign separate tokens for different visit types, such as start/end tokens for inpatient, outpatient, and emergency visits. Each visit creates a temporal segment, so we place a [SEP] token before the start of each visit. We also calculate age tokens (ranging from 0 to 119) based on the visit date and include them with the visit tokens, enabling the model to track long-term disease progression where age is an important factor.

For laboratory measurements containing continuous numerical values, we create 10 quantiles for each measurement type and assign quantile tokens ([Q1] through [Q10]) to represent the discretized values. This quantile-based approach, adapted from Ethos~\cite{renc_zero_2024}, allows us to effectively utilize numerical laboratory values within the discrete token framework.

\subsection{Training}
We adopt the standard Masked Language Model (MLM) method\cite{Devlin2019BERTPO} to pretrain our model. To finetune the model for the downstream task of predicting chronic kidney disease prognosis, we set binary classification objectives across several follow-up periods. We define case/control groups for each task with different follow-up and assessment period combinations. We then finetune the model to predict whether a patient will progress to a specific CKD stage within the designated follow-up period. For example, one task predicts whether a patient will progress from stage 3a to stage 5 within 365 days.

%% file: 3_experiments.tex
\section{Experiments}
\label{sec:experiments}

\subsection{Experimental Setup}
We implemented our model using PyTorch and trained it on an A6000 GPU. The model consists of 6 transformer layers with 6 attention heads and a hidden dimension of 768. We used Adam optimizer with learning rate of 1e-4 and trained for 64 epochs with batch size of 32.

\subsection{Baseline Model}
We compared our model against CEHR-BERT \cite{pang_cehr-bert_nodate},  a prominent and well-established benchmark within the field of electronic health record analysis. CEHR-BERT employs [masked language modeling on EHR sequences] and has demonstrated strong performance on various clinical prediction tasks. It utilizes demographic, drug, diagnosis, and procedure information.

\subsection{Evaluation Metrics}
We evaluated model performance using multiple metrics: Receiver Operating Characteristic-Area Under the Curve (ROC-AUC) and Precision-Recall-Area Under the Curve (PR-AUC) for discrimination ability, and Accuracy, Precision, Recall, and F1 score for classification performance. PR-AUC is particularly important given the class imbalance inherent in CKD progression prediction.

\subsection{Evaluation Scenarios}
To comprehensively evaluate predictive capability, we designed experiments across multiple temporal dimensions:
\begin{enumerate}
    \vspace{0.1cm}
    \item \textbf{Follow-up Period Variation}
    \\ 180, 365, 730, 1095, and 1460 days
    \vspace{0.1cm}
    \item \textbf{Assessment Period Variation}
    \\ 180, 365, and 730 days
\end{enumerate}
\vspace{0.1cm}
The experimental design encompassed all 15 combinations of the follow-up and assessment periods described above. In the subsequent results section, performance metrics for each specific follow-up period and assessment period are reported as averaged values.

%% file: 4_results.tex
\section{Results}
\label{sec:results}

\begin{table*}[!htp]
\centering
\small
\begin{tabular}{>{\centering\arraybackslash}p{3.5cm}lllllllll}
\hline
Follow-up Period (days) & Model     & ROC-AUC & PR-AUC & Accuracy & Specificity & Precision & Recall & F1     \\ \hline
\multirow{2}{*}{180}  & CEHR-BERT & 0.8729  & 0.8224 & 0.7806   & 0.8556      & 0.8042    & 0.7037 & 0.7415 \\
                      & \textbf{ProQ-BERT(Ours)}     & 0.9949  & 0.9889 & 0.9630   & 0.9744      & 0.9444    & 0.9444 & \textbf{0.9394} \\ \hline
\multirow{2}{*}{365}  & CEHR-BERT & 0.8357  & 0.8404 & 0.7032   & 0.8029      & 0.7389    & 0.5983 & 0.6547 \\
                      & \textbf{ProQ-BERT(Ours)}      & 0.9970  & 0.9940 & 0.9625   & 0.9710      & 0.9167    & 0.9524 & \textbf{0.9267} \\ \hline
\multirow{2}{*}{730}  & CEHR-BERT & 0.8630  & 0.8582 & 0.7757   & 0.7815      & 0.7763    & 0.7708 & 0.7704 \\
                      & \textbf{ProQ-BERT(Ours)}      & 0.9418  & 0.8754 & 0.9058   & 0.9319      & 0.8342    & 0.8586 & \textbf{0.8413} \\ \hline
\multirow{2}{*}{1095} & CEHR-BERT & 0.8222  & 0.8031 & 0.7672   & 0.8488      & 0.8151    & 0.6845 & 0.7404 \\
                      & \textbf{ProQ-BERT(Ours)}      & 0.9610  & 0.9301 & 0.9174   & 0.9717      & 0.9190    & 0.7894 & \textbf{0.8483} \\ \hline
\multirow{2}{*}{1460} & CEHR-BERT & 0.8374  & 0.8127 & 0.7710   & 0.8189      & 0.8007    & 0.7229 & 0.7524 \\
                      & \textbf{ProQ-BERT(Ours)}      & 0.9543  & 0.9148 & 0.8890   & 0.9532      & 0.8444    & 0.7184 & \textbf{0.7735} \\ \hline
\end{tabular}
\caption{Performance comparison across different follow-up periods}
\label{tab:followup_results}
\end{table*}

\begin{table*}[!htp]
\centering
\small
\begin{tabular}{>{\centering\arraybackslash}p{3.5cm}lllllllll}
\hline
Assessment Period (days)          & Model     & ROC-AUC & PR-AUC & Accuracy & Specificity & Precision & Recall & F1     \\ \hline
\multirow{2}{*}{180} & CEHR-BERT & 0.8302  & 0.8385 & 0.7518   & 0.8107      & 0.7833    & 0.6904 & 0.7289 \\
                     & \textbf{ProQ-BERT(Ours)}      & 0.9694  & 0.9511 & 0.9148   & 0.9519      & 0.8569    & 0.8333 & \textbf{0.8385} \\ \hline
\multirow{2}{*}{365} & CEHR-BERT & 0.8180  & 0.8015 & 0.6940   & 0.8362      & 0.7645    & 0.5482 & 0.6366 \\
                     & \textbf{ProQ-BERT(Ours)}      & 0.9662  & 0.9425 & 0.9298   & 0.9535      & 0.8805    & 0.8674 & \textbf{0.8668} \\ \hline
\multirow{2}{*}{730} & CEHR-BERT & 0.8905  & 0.8421 & 0.8328   & 0.8177      & 0.8133    & 0.8495 & 0.8302 \\
                     & \textbf{ProQ-BERT(Ours)}      & 0.9738  & 0.9283 & 0.9380   & 0.9759      & 0.9378    & 0.8572 & \textbf{0.8923} \\ \hline
\end{tabular}
\caption{Performance comparison across different assessment periods}
\label{tab:assessment_results}
\end{table*}

\subsection{Overall Performance}
Tables \ref{tab:followup_results} and \ref{tab:assessment_results} present the performance comparison between ProQ-BERT and CEHR-BERT. Our model consistently outperforms CEHR-BERT across all metrics and temporal settings.

\subsection{Follow-up Period Analysis}
Table \ref{tab:followup_results} shows performance across different follow-up periods (180-1460 days). For short-term prediction (180 days), our model achieves an ROC-AUC of 0.995 and PR-AUC of 0.989, substantially outperforming CEHR-BERT (0.873 and 0.822). Even for long-term prediction (1460 days), our model maintains strong performance (ROC-AUC: 0.954, PR-AUC: 0.915) compared to CEHR-BERT (0.837 and 0.813).

The performance gap widens as the follow-up period decreases, with the largest difference at 180 days (ROC-AUC difference: 0.122), suggesting our model effectively captures early markers of rapid disease progression.

\subsection{Assessment Period Analysis}
Table \ref{tab:assessment_results} presents performance across varying historical data lengths. Our model achieves strong performance with an assessment period of 180 days (ROC-AUC: 0.969, PR-AUC: 0.951). Performance remains consistently high across all periods, with the longest period (730 days) yielding ROC-AUC of 0.974 and PR-AUC of 0.928. The consistent performance across different assessment periods suggests that meaningful predictions can be made from varying amounts of historical data.

%% file: 5_conclusion.tex
\section{Conclusion}
\label{sec:conclusion}

Our \textbf{ProQ-BERT} demonstrated a substantial improvement over the baseline approach, highlighting the effectiveness of incorporating transformer-based architectures and laboratory data representations. Furthermore, experiments across varying assessment and follow-up periods revealed two key trends: (1) performance decreased as the follow-up horizon lengthened, likely due to the increased uncertainty in predicting distant outcomes; and (2) extending the assessment period improved predictive accuracy, suggesting that a broader temporal context provides more informative patterns for prognosis estimation. These findings underscore the importance of temporal design choices in clinical prediction models and offer guidance for future research in optimizing time-window configurations.